# Shift Convolution Network for Stereo Matching

Jian Xie

Hefei University of Technology

## Abstract

*In this paper, we present Shift Convolution Network (ShiftConvNet) to provide matching capability between two feature maps for stereo estimation. The proposed method can speedily produce a highly accurate disparity map from stereo images. A module called shift convolution layer is proposed to replace the traditional correlation layer to perform patch comparisons between two feature maps. By using a novel architecture of convolutional network to learn the matching process, ShiftConvNet can produce better results than DispNet-C[1], also running faster with 5 fps. Moreover, with a proposed auto shift convolution refine part, further improvement is obtained. The proposed approach was evaluated on FlyingThings 3D. It achieves state-of-the-art results on the benchmark dataset. Codes will be made available at github.*

1. Introduction

Depth estimation through a stereo pair of images is a fundamental problem in the field of computer vision. It aims at predicting per pixel disparity by estimating correspondences between the left and right images. It has a number of applications in practice, e.g., autonomous vehicles, robotics navigation, augmented realities (AR). Moreover it is also used to support other computer vision tasks including recognition and 3D reconstruct.

Researchers has been working on depth estimation for several decades. Conventional methods contain four steps[2]: step 1. matching cost computation; step 2. cost support aggregation; step 3. disparity computation, and step 4. disparity refinement. Step 1 to 3 are often considered as a global optimization problem for choosing a best disparity, which can lead to minimal matching cost for each pixel.

Recently, taking advantage of deep fully convolution networks, many works adopt end-to-end pipeline to yield high-quality disparity images. A pair of rectified images are directly fed into the pipeline to get depth values for each pixel. These methods can be divided into two categories in terms of whether or not consisting an extra process to compute matching cost. Networks without the extra process[3, 4] uses concatenated left and right images on the color channel as the input of the network, passing through an encode part with several convolution layers to finish step 1 and step 2 together. Due to the limited size of kernels in convolution, it's difficult to detect stereo correspondence when there exist a huge displacement between some patches. In order to associate patches in different regions to generate better matching clues, most methods[1, 5-7] apply an extra process to compute matching cost. To do that, they typically adopt a correlation layer. It is a hand-crafted process which use left and right feature maps as inputs. The key part is shifting, which is supposed to move corresponding regions on two feature maps to the same position. Then multi-scale multiplicative patch comparisons can be produced by shifting feature maps at different scales. Since simply applying multiplication between feature maps has limited capability on matching computation, the results are not always promising.

In order to get better patch comparisons automatically, we proposed a simple but efficient network called ShiftConvNet to learn the correlation of the left and right feature maps. Basically, we propose a shift convolution layer to produce matching cost volume, which pass forward to one encode part and several up-sampling blocks as decode part to obtain a full-size disparity map. Moreover we designed an auto shift convolution refine part to further improve the accuracy by adding patch comparisons on original images.

Our model was trained end-to-end on Flying Things 3D dataset at a single GPU machine for one or two days. Our networks achieved state-of-the-art accuracy and also remained high speed.

In addition, we also replaced the correlation layer in DispNet-C with our proposed shift convolution layer in order to proof its effectiveness. Experimental results have shown shift convolution layer improved the accuracy of the method.

The contributions of this paper are:

(1). A unified framework called ShiftConvNet for stereo matching, which achieved state-of-the-art accuracy on the FlyingThings 3D.

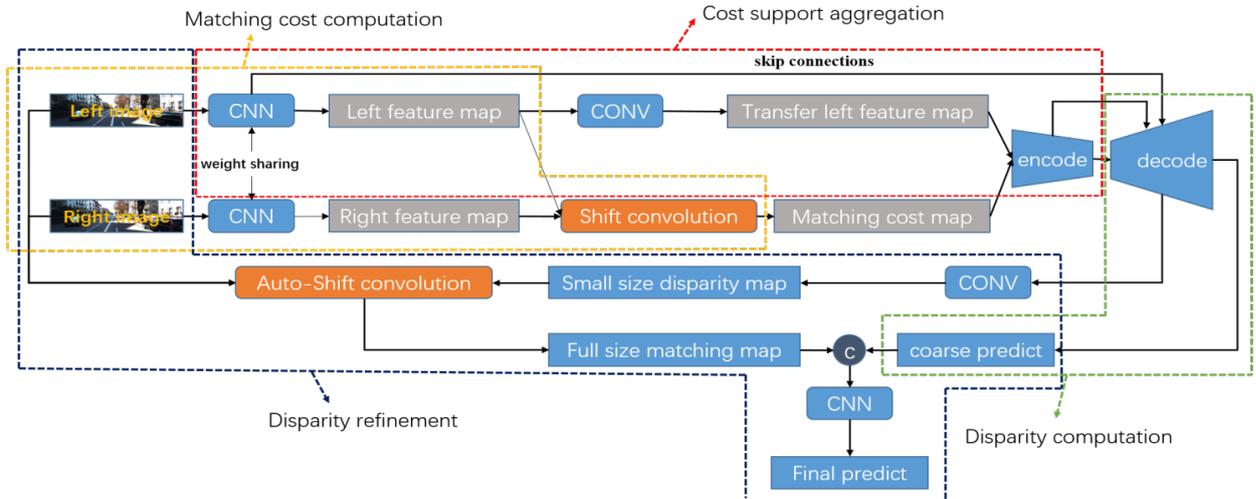

Figure 1. Architecture of our proposed ShiftConvNet. Left and right stereo images are fed into shared-weight CNN for feature extraction. Left feature map and right feature map are used for matching volume computation. An intermediate output from the decode part is used to generate small size disparity map, based on which sift range is dynamically decided for Auto_shift convolution.

(2). A new structure for matching cue computation. Compared to widely used correlation layer, it can produce better patch corresponding information between feature maps.

(3). A SGM (Semi-Global Matching) refine part called auto-displacement refine part which use the mode numbers of a small-size coarse disparity map to guide shifting process.

2. Related Work

**Fully convolutional neural network (FCN).** Fully convolutional neural network was firstly used to solve semantic segmentation problem. In recent years, it has been widely used in pixel-wise labeling tasks, including semantic segmentation[8, 9], optical flow[10, 11] and also depth estimation. Fully convolutional neural network with only convolution layers allows to generate per pixel prediction. FlowNet[12] first used fully convolution network with end-to-end training on optical flow. Recent stereo estimation methods mostly rely on FCN and gain a lot improvement.

**Methods without indicated process for matching cost computation.** With fully convolutional neural network widely used in disparity estimation, a number of methods simply use several convolution layers to accomplish both step 1 (matching cost computation) and step 2 (cost support aggregation) without an explicit process for step 2. For example, DispNet-S[1] with concatenated left and right image on color channels as the input, use an encode part with several convolution layers to implement cost computation and cost support aggregation together. Some other methods[3] focusing on disparity map refinement pay more attention to how to apply semi-global matching (SGM)[13], often without an indicated step for matching cost computation. CSPN[3] performed well with a simplified sequential network to compute disparity and a convolutional spatial propagation network to refine disparity.

**Methods with indicated process for matching cost computation.** Traditional methods[14, 15] design local descriptors to fulfil matching cost computation. According to [2], the ordinary pixel-based matching cost including squared intensity difference[16-19] and absolute intensity difference[20]. Other conventional approaches for matching cost computation include normalized cross correlation[17, 21] and binary matching cost[22]. CNN is firstly used for matching cost computation by Zbontar and LeCun[23]. In order to compute matching clues for optical flow, Dosovitskiy Alexey et al.[10] proposed a correlation layer, which is a matching cost computation module. It performs multiplicative patch comparisons between two feature maps. In the first version of the correlation layer, shifting process is done on two dimensions and used on optical flow. Mayer et al.[1] put forward DispNet-C with a 1D-correlation layer by changing the shifting process to only a horizontal dimension to achieve high speed. Rather than picking the matching patches, matching clues generated by correlation layer are added into the network. After that, 1D-correlation layer has been extensively used for matching cost computation in stereo estimation, including DESNet[5], which apply it on two different feature scales via multi-scale feature extraction, with smaller displacement on shallow feature maps and larger displacement on deep feature maps. Other methods with the correlation layer adding extra information for disparity aggregation, e.g. semantic information[6], edge information[24]. Other methods like PSM-Net[7], features from a SPP module was adopted to form cost volume by concatenated left feature maps with right feature maps across every disparity level.

## 3. Approach

We present ShiftConvNet, different from exiting methods, ShiftConvNet using a proposed shift convolution layer to encode matching clues. We connect the network with a proposed refine part called auto_disp_refine. With shift convolution layer and taking advantage of auto_disp_refine our approach can yield state-of-the-art results. In this part we firstly introduce our framework and the shift convolution layer is introduced next, finally auto_disp_refine is presented.

### 3.1. Network structure

The framework of our proposed ShiftConvNet is shown in figure 1. Our objective is to predict an accurate disparity map. Like most recent end-to-end stereo matching systems, we incorporate all four steps into ShiftConvNet. It has an encode part which severs as cost support aggregation with the shift convolution layer to compute matching cost, a decode part to compute disparity, and an auto shift refine module to optimize the result.

As illustrated in Figure 1, the left and right images are first separately fed into shared-weight CNNs to fulfil feature extraction to get left feature maps and right feature maps. The CNN in this part contains four convolution layers and two pooling layers. Unlike in other studies using large kernel sizes, such as 7 in the first and 5 in the second convolution layer in DispNet-C[1], inspired by VGG[25], ShiftConvNet only use small filters (3*3), but go deeper to remain the same receptive filed. The left and right feature maps then pass forward into the shift convolution layer to generate patch comparison information. At the same time, left feature maps are fed into a convolution layer to get transferred left feature maps (denoted as conv_redir in Figure 2). Then the matching cost maps and transferred left feature maps are concatenated and fed into the encode block to finish the cost support aggregation process. The encode block consists of several convolution layers and pooling layers, from conv5 to conv8 illustrated in Figure 2. For disparity computation, the aggregated high-level feature maps pass through the decode part, which comprises 6 upsampling blocks. Each one includes a deconvolution layer for upscaling and a convolution layer for smoothing. To guide disparity computation there are several skip connections from "cost support aggregation" part to the decode part. The skip connections from the CNN of feature extraction to the decode part are only applied to the subnet for the left image, since the reference image is the right image, which is only used to compute matching

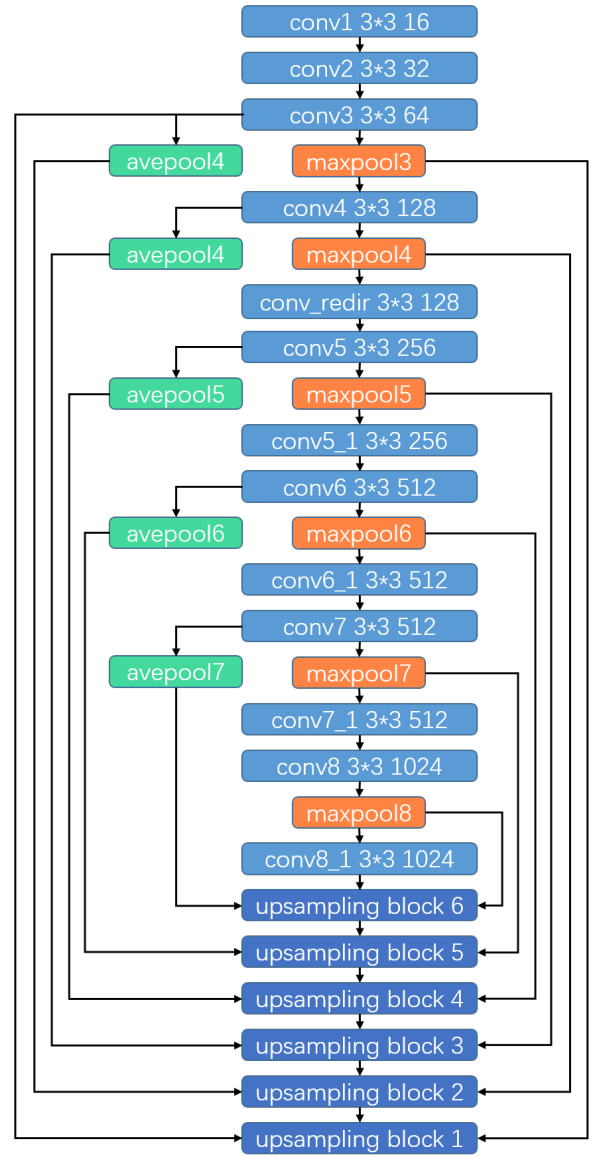

Figure 2 The details of the "cost support aggregation" part, decode part and the skip connections between them

cues for cost support aggregation. The goal of our network is to compute the left disparity map.

The details of the skip connections are presented in figure 2. Skip connections link the pooling layers in encode part to the deconvolution layers in upsampling blocks. The reason for that is to add significant features to guide upscaling. And feature maps from feature extraction part are concatenate with upscaled feature maps to provide background information for smoothing.

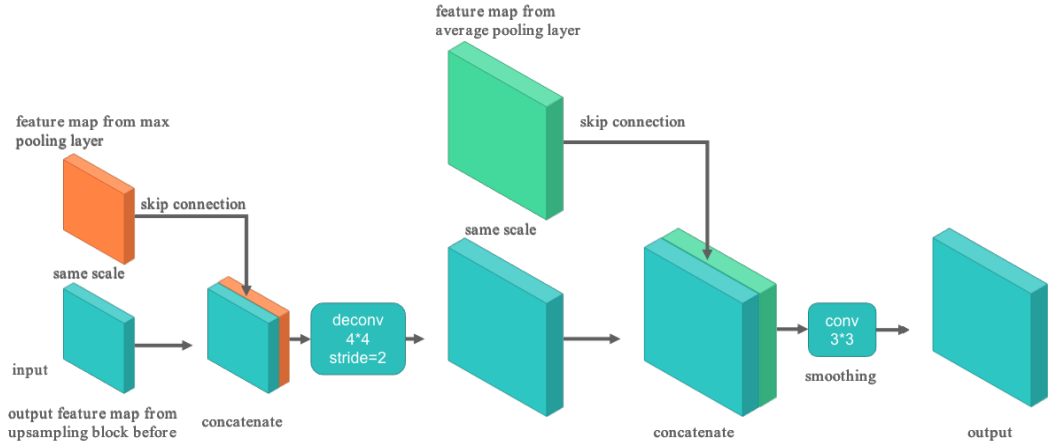

Figure 3 Architecture of an upsampling block.

The structure of the upsampling block is illustrated in Figure 3. Every block has two skip connections, which come from the output feature maps of the corresponding layer of the feature extraction CNN part and the corresponding pooling layer of the encode part. The linked feature maps with the same size are concatenated to complete the upscaling and smoothing.

### 3.2. Shift convolution layer

In stereo estimation community, the correlation layer is supposed to be a fundamental structure to compute matching cost. Although it performs well in many cases, it doesn't have the ability of learning. It uses a multiplication of shifted feature maps to compute patch corresponding. In order to learn producing better patch comparisons, we create a new process called shift convolution layer, which comprises three steps: shift concatenation, matching clue computation, and concatenation. The module is illustrated in figure 4.

**Shift concatenation.** The motivation of this step is to put all pairs of patches under all reasonable shift values to the same position, so the convolution with small kernel size, e.g. (3,3), can be used to compute the matching cost between two patches where there exist a displacement. To achieve this, left feature maps are firstly sliced along the horizontal dimension. The start point of the slicing, indicated as displacement, is changing from 0 to "maxdisp" (max displacement of feature map, setting to 40 in our experiments) along the left-to-right direction. The end point of the slicing is fixed to the rightmost of the left feature maps. Then the resulting slices of feature maps are padded with zeros on the right side to keep their sizes unchanged. After that, these maps are concatenated with right feature maps. At the same time, the analogous operations are done to the right feature maps, with the starting point of the slice ranging from the right side to "maxdisp", the end point fixed to the left side, and with zeros padding on the left side.

Then they are concatenated with left feature maps.

**Matching clue computation.** After each scale of shift concatenation, the result feature maps pass forward to a convolution layer to learn how to compare pairs of feature patches. In this article we tried 8, 12, 16 filters with kernel size (3, 3) in this convolution layer to show an increase number of filters can improve its performance.

**Concatenation.** After matching cost computation from different shifting scales, shift convolution layer combine these matching cost maps by concatenating them to form the total matching cost volume as the output of this module.

### 3.3. Auto_disp_refine part

We propose a refine structure using auto shift convolution layer to leverage patch comparison information on the original images to improve accuracy of final results. The shift convolution layer is used to compute patch comparison directly on a pair of original images, as illustrated in Fig. 1. Because of large disparity between original full-size left and right images (sometimes more than 200), shift processing on original images need to range large scales to make sense, which would hugely increase the time cost. So rather than using a constant shifting range, the refine structure uses a small disparity map to guide the shifting process. The small disparity map is generated by a convolution layer which receives the output of the upsampling block 6 of the decode part as an input (see Fig. 1 and Fig. 2). Disparities on this small disparity map are considered as 'mode numbers' of the ground truth. We chose the scales within error range $\pm 2$ from the small disparity map as shifting scales. With specific scales of shifting, patch comparison information on the original image pair is obtained by the proposed shift convolution (because the displacement is auto decided, so here we call it auto-shift convolution). The results of all shifting scales are simply added together to get the final matching map. After that they concatenate with the coarse disparity map outputted from the decode part. Then the concatenated maps are inputted into a CNN, which have three convolution layers, to get the final refined disparity map. The numbers of filters of the three convolution layers are 16, 32, and 1. All filters in our architecture have a small size of 3*3.

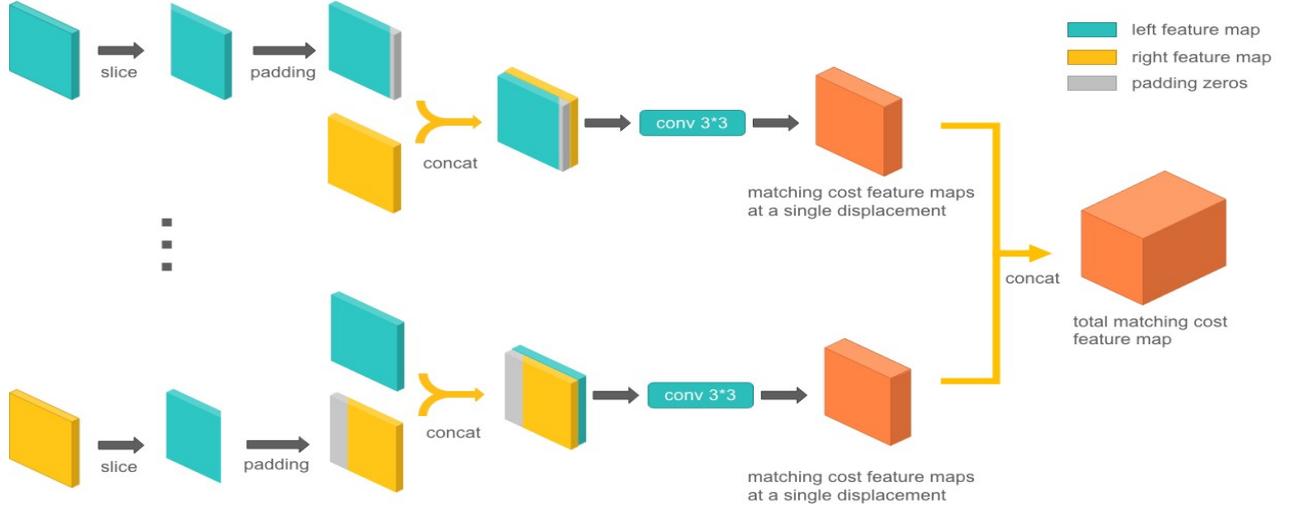

Figure 4. Framework of shift convolution layer. Gray parts of feature maps indicate regions for padding, feature maps generated by convolution for each scale of slice and padding are concatenated together to form total matching cost feature map as the output of the whole structure.

3.4. Training

For training, we construct two loss functions as follows:

$$Loss_1 = L_{1smooth}(p_c, T) + \alpha L_2(w)$$
$$Loss_2 = L_{1smooth}(p_f, T) + \alpha L_1(p_s, T_s) + \beta L_2(w)$$

Where $p_c$, $p_s$, $p_f$ indicate coarse prediction, small size prediction, final prediction. $T$ indicates ground truth, and $T_s$ indicates small size ground truth, which is acquired by resizing the ground truth using nearest neighbor interpolation. $\alpha$, $\beta$ are hyper parameters. $w$ denotes weights. $L_1$ is Manhattan Distance, $L_2$ Euclidean distance, and $L_{1smooth}$ is the smooth function applied on the $L_1$ distance as below:

$$smooth_{L_1}(x) = \begin{cases} 0.5x^2, & \text{in where } |x| < 1 \\ |x| - 0.5, & \text{others} \end{cases}$$

We firstly train network without refine part using $Loss_1$, and then train with refine part using $Loss_2$.

4. **Experiments**

We evaluated our network on FlyingThings 3D subset[1]. FlyingThings 3D subset is a large synthetic dataset omitted some extremely hard samples from FlyingThings 3D, which contains 21,818 pairs of training and 4,248 pairs of testing images with full density disparity map for each image as ground truth. The size of the images is width=960, height= 540. On training we resize images and disparity to (384*768).

We also performed ablation experiments on FlyingThings 3D subset by changing the number of filters of the convolution layer in shift convolution layer. Moreover, in order to validate the practicability of the proposed shift-convolution-layer, we designed ablation studies by using shift convolution layer as a replace of correlation layer in DispNet-C[1]. The implementation details present in subsection 4.1, and the experimental results on the FlyingThings 3D in subsection 4.2.

4.1. Implementation Details

The ShiftConvNet architecture was implemented using TensorFlow. Losses during training show in figure 5 supported by TensorBoard. During training, images were resized to H =384, W=768 using nearest neighbor interpolation. On FlyingThings 3D subset, we set learning rate to 0.0002 for the first 100,000 iterations, and divided by 2 for each 50,000 iterations to reach the lower bound 0.00003 for the rest of iterations. Training process was running on an Nvidia GTX1080 GPU for approximately 50 hours.

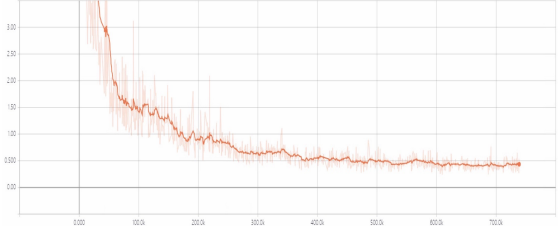

Figure 5. The losses during training

## 4.2. Results on Flyingthings 3D

On Flyingthings 3D, we designed several ablation experiments. The ShiftConvNet with the best setting yielded a 1.10 end-point error.

**Ablation study for shift convolution layer.**

In order to compare two slightly different designs of the shift convolution layer, we experimented placing the convolution layer after concatenating all scales of displacements. We also analyses the influence of the number of filters of the convolution layer.

As listed in Table 1, for shift convolution layer, placing the convolution layer before concatenating the maps of different displacements has shown better results and higher speed than convolution after concatenating. This is because concatenating first increases parameters in the feature extraction convolution. And integrating feature maps from all displacements seems to confuse the convolution layer, so it is more difficult to learn matching. Therefore in the next experiments, we used the architecture presented in Fig. 4.

The results in Table 1 also show more filters can improve final results but also increase the time cost. The results in the bottom row of Table 1 is obtained with 16 filters in the convolution layer of shift convolution layer and the proposed refinement method.

Table 1 The results (End Point Error, EPE) of ablation experiments on Flyingthings 3D validation set

|  | # of Matching clue computation filters | Running time | EPE |
|---|---|---|---|
| Feature extraction first (No refine) | 8 | 0.22s | 1.47 |
|  | 12 | 0.24s | 1.41 |
|  | 16 | 0.26s | 1.44 |
| Concatenate first (No refine) | 8 | 0.21s | 1.33 |
|  | 12 | 0.22s | 1.26 |
|  | 16 | 0.23s | 1.15 |
| with refine | 16 | 0.31s | 1.10 |

**Comparisons with correlation layer.**

To demonstrate that shift convolution layer has the ability to learn patch comparison and can produce better results than correlation layer, we replaced correlation layer with shift convolution layer on DispNet-C After using shift convolution layer, the EPE of DispNet-C improved from 1.67 to 1.46, which demonstrated the capability of our structure.

**Comparisons with other state-of-the-art methods.**

We compared our method with other state-of-the-art methods, including Dispnet-C[1], PSMNet[7], SegStereo[6], and SGM[13]. As shown in Table 2, ShiftConvNet perform more accurate than other methods and even better in terms of running speed. Figure 6 illustrates some examples of the disparity maps estimated by our ShiftConvNet.

Table 2 Comparisons of our method with other state-of-the-art methods on Flyingthings 3D validation set

| Methods | EPE (End Point Error) | D1(3 pixel error) |
|---|---|---|
| ShiftConvNet | 1.10 | 5.56 |
| Dispnet-C[1] | 2.33 | 10.04 |
| PSMNet[7] | 1.32 | 6.20 |
| SegStereo[6] | 1.45 | 3.50 |
| SGM[13] | 7.26 | 16.18 |

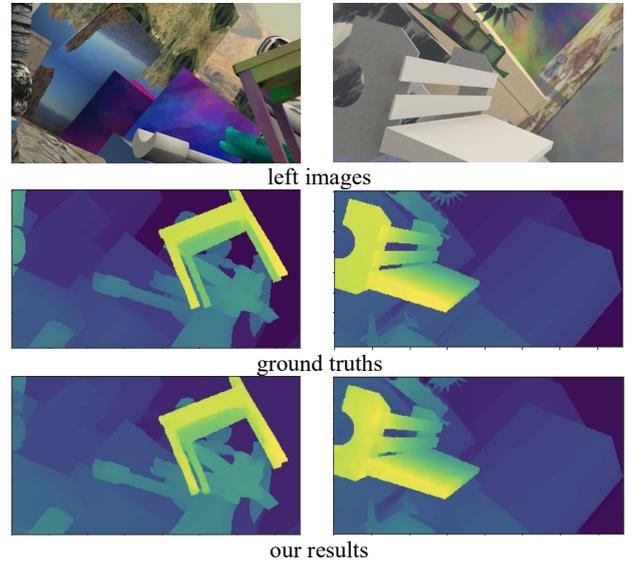

left images

ground truths

our results

Figure 6. Qualitative examples of our method

## 5. Conclusions

In this paper we proposed a new framework for stereo estimation which contains a novel patch comparison structure that performed better then correlation layer and a refinement part which can automatically choose shift value to generate image patch matching cost for disparity refinement. Experiments on FlyingThings 3D subset demonstrated the effectiveness of our framework. Ablation experiments for shift convolution layer on FlyingThings 3D have proved that with shift convolution layer substituting correlation layer, the benchmark method can produce more accurate disparity map. Our network achieved state-of-the-art performance on FlyingThings 3D dataset. Qualitative evaluation on FlyingThings 3D further manifested its capability.